\documentclass[conference]{IEEEtran}  
\IEEEoverridecommandlockouts

\usepackage{flushend}
\usepackage{multirow}
\usepackage{adjustbox}
\usepackage{cite}
\usepackage{amsmath,amssymb,amsfonts}
\usepackage{algorithmic}
\usepackage{graphicx}
\usepackage{textcomp}
\usepackage{xcolor}
\usepackage{siunitx}
\def\BibTeX{{\rm B\kern-.05em{\sc i\kern-.025em b}\kern-.08em
    T\kern-.1667em\lower.7ex\hbox{E}\kern-.125emX}}

\begin{document}

\title{	Automatic occlusion removal from 3D maps for maritime situational awareness}

\author{
	\IEEEauthorblockN{Felix Sattler\IEEEauthorrefmark{1}, Borja Carrillo Perez\IEEEauthorrefmark{1}, Maurice Stephan\IEEEauthorrefmark{1}, Sarah Barnes\IEEEauthorrefmark{1}}
	\IEEEauthorblockA{\IEEEauthorrefmark{1}German Aerospace Center (DLR), Institute for the Protection of Maritime Infrastructures,
		Bremerhaven, Germany}
}

\maketitle

\begin{abstract}
We introduce a novel method for updating 3D geospatial models, specifically targeting occlusion removal in large-scale maritime environments.
Traditional 3D reconstruction techniques often face problems with dynamic objects, like cars or vessels, that obscure the true environment, leading to inaccurate models or requiring extensive manual editing.
Our approach leverages deep learning techniques, including instance segmentation and generative inpainting, to directly modify both the texture and geometry of 3D meshes without the need for costly reprocessing.
By selectively targeting occluding objects and preserving static elements, the method enhances both geometric and visual accuracy.
This approach not only preserves structural and textural details of map data but also maintains compatibility with current geospatial standards, ensuring robust performance across diverse datasets.
The results demonstrate significant improvements in 3D model fidelity, making this method highly applicable for maritime situational awareness and the dynamic display of auxiliary information.
\end{abstract}


\section{INTRODUCTION}
\label{sec:intro}  

\noindent In the context of maritime security, different stakeholders such as port authorities, law enforcement agencies and research institutions maintain large, geospatial 3D assets (for example, digital surface models, DSM) that are used for situational awareness and on-site monitoring.
Static 3D information is used as a geospatial layer onto which different auxiliary information is displayed dynamically \cite{wieland2023}.
When performing 3D reconstruction of static maritime environments from remote sensing data, dynamic objects that occlude the environment (occluders) are almost always present.
In port infrastructures this can refer to berthed vessels on water bodies, shipping containers in terminals or parked vehicles along the quay.
Generating a 3D map that incorporates these occluding objects does not reflect the true environment and limits the insertion of auxiliary information into the 3D map.
The generation of large 3D assets is resource intensive and requires specialized processing techniques such as photogrammetry which is capable of producing 3D geometries from remote sensing data.

Simply removing all occluding objects manually from the collected imagery presents two disadvantages:
First, masking out objects during preprocessing by retouching further increases resource demand during model generation.
Additionally, especially in large regions with sparse pattern information (for example water bodies or container depots) occluders provide important features that help to register images more robustly and thus improve the reconstructed geometry.
Therefore, their removal during preprocessing is not always feasible.
In this work, we present a novel method for direct 3D geometry processing to enable the removal of occluders as a postprocessing step.
With this method, existing 3D assets can be reprocessed to enable the insertion of auxiliary information.
Users, such as scientific staff, government authorities or analysts can reuse existing DSMs instead of creating new ones.
Our framework combines state-of-the-art instance segmentation and generative inpainting using deep learning with projection mapping to correct surface textures and remesh geometry information.
The proposed method is robust and allows users to select classes of occluding objects that will be removed automatically without regenerating the whole 3D asset. 
In the remainder of this paper an overview of related works will be given (Section \ref{sec:relworks}), then the method will be introduced (Section \ref{sec:method}), followed by an application to a real-world dataset (Section \ref{sec:results}) and a summary (Section \ref{sec:conclusion}).

\begin{figure*} [!h]
   \includegraphics[width=\textwidth]{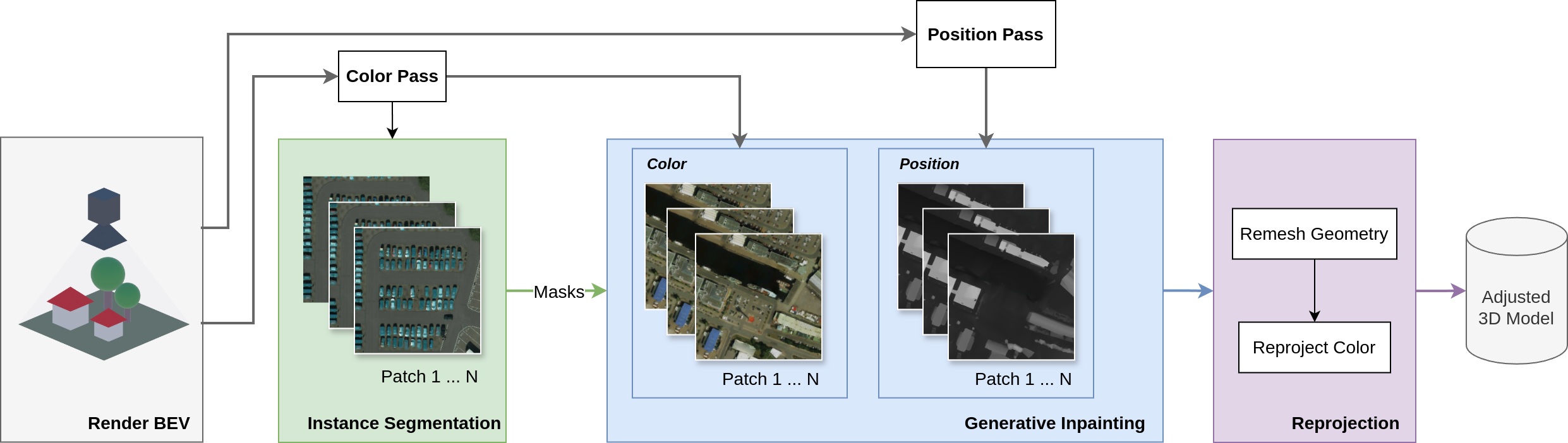}
   \caption[pipeline] 
   { \label{fig:pipeline} 
    Overview of the proposed method for updating 3D geospatial models using mask-aware inpainting and geometric remeshing. First, an orthogonal view (bird's eye view, BEV) of the map is renderer in patches. Then, user-defined classes of objects are detected using instance segmentation. The resulting masks are used to control mask-aware inpainting of color and 3D position passes. The final step involves remeshing the geometry using the 3D position pass and reprojecting the inpainted color data onto the 3D model, resulting in an accurately adjusted representation of the environment.}
\end{figure*} 

\section{RELATED WORKS}
\label{sec:relworks}

To effectively modify a 3D mesh, it is essential to alter both texture and geometry information.
A key technique utilized in this context is mask-aware inpainting, which has become increasingly significant in recent advancements in image processing \cite{surveyinpainting}.
Inpainting methods aim to fill or reconstruct missing or occluded regions of an image or 3D model, guided by a mask that specifies the areas to be reconstructed.
Traditionally, inpainting methods relied on patch-based\cite{criminisi2004} or geometric constraints\cite{huang2014}, which often struggle with complex details and large occlusions, leading to artifacts or blurring\cite{imagemelding2012}.
Recent deep learning methods, such as those based on generative adversarial networks (GANs)\cite{zhao2021comodgan}, diffusion models\cite{repaint2022}, and image convolutions\cite{suvorov2021resolution} in the frequency domain, have significantly improved inpainting capabilities.
\newline In remote sensing, 2D inpainting has been used successfully in various applications.
For instance, GANs have been employed to fill in cloud-occluded areas in satellite images\cite{kuznetsov2020, kumar2022gan}, where traditional methods fall short.
Similarly, inpainting techniques have been applied to remove vehicle occlusions to produce consistent lane markings for semantic analysis\cite{seyed2023}, and to reconstruct false-color data in sea surface temperature (SST) images\cite{sst2019}.  
This demonstrates the versatility of inpainting across different imaging modalities.

2D inpainting techniques have also been applied to alter 3D data.  
Engels \textit{et al.} \cite{engels2011facades} combined traditional inpainting with plane-fitting to modify 3D point clouds of building facades in urban scenes, though the method is limited by poor mask detection and inpainting quality.  
To improve upon this, recent approaches have proposed using state-of-the-art 2D inpainting methods to remove objects from images and then refit the 3D model with the new data.
Due to the ability of deep-learning inpainting methods to generalize across a variety of imaging data, it is possible to modify non-color data such as depth information or 3D position data.
Mirzaei \textit{et al.}\cite{mirzaei2023} proposed a method called SPIn-NeRF generating a neural radiance field (NERF)\cite{mildenhall2021} and reoptimizing it with 2D inpainted depth and color data.
Similarly, Prabhu \textit{et al.} \cite{prabhu2023inpaint} jointly optimized a NERF and a diffusion model for 2D inpainting.

However, these methods encode data as neural representations that are difficult to integrate with modern geospatial data standards like 3DTiles OGC\cite{3dtiles}.
Also, a reoptimization of neural representations is slow and requires users to convert existing 3D data into a suitable format.
Nevertheless, works like SPIn-NeRF  showcase the potential of inpainting for optimizing 3D geometry.
Building on these advances, we introduce a novel method that combines 2D instance segmentation and mask-aware inpainting with 3D reprojection and remeshing.
We perform instance segmentation on an orthogonal bird's eye view of the 3D map and then apply inpainting to color and elevation data.
This approach directly modifies the surface texture and geometry of 3D meshes without the need for expensive retraining or recomputation.
This approach allows users to reuse existing 3D data, such as DSMs and is conceptualized to work robustly with large 3D datasets.

\section{PROPOSED ARCHITECTURE}
\label{sec:method}

\begin{figure*} [ht]
   \includegraphics[width=\textwidth]{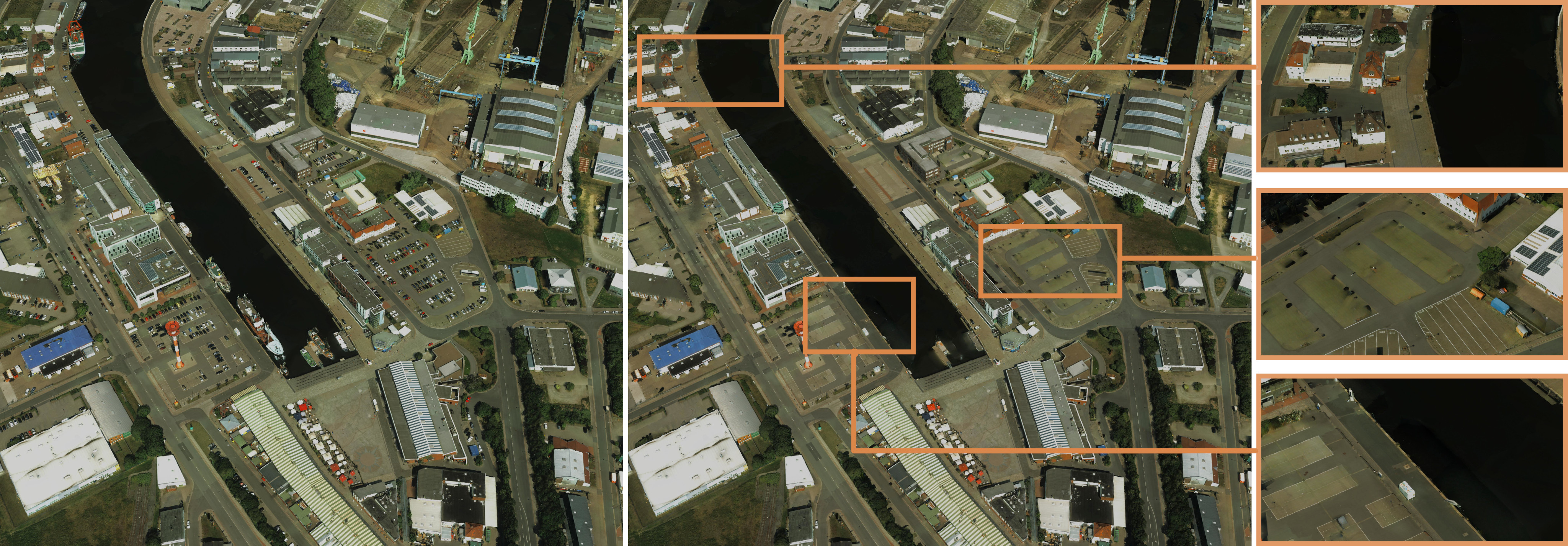}
   \caption[collageprepost] 
   { \label{fig:collage_pre_post} 
   A qualitative comparison of our framework applied to a 3D geospatial map inpainted using LaMa \cite{suvorov2021resolution}. The left side depicts the original 3D scene, while the middle illustrates the results after inpainting and remeshing. Occluded areas and dynamic features have been seamlessly reconstructed and updated. The highlighted regions on the far right demonstrate the fidelity of our technique, which effectively preserves structural details and ensures consistent and accurate updates to the geospatial data.}
\end{figure*} 

Our method for updating 3D geospatial models integrates mask-aware inpainting and geometric remeshing, optimized for remote sensing data from satellite or aerial applications.
In Figure \ref{fig:pipeline} the process is illustrated.
It begins by setting up an orthogonal camera in a bird's-eye view (BEV).
This camera setup ensures that the projection is consistent across the entire scene, avoiding perspective distortions.
The camera transformation matrix and projection matrix from BEV to 3D mesh are stored for later use in remeshing and reprojection.
Then, a color (RGB) and a height map are rendered.
The height map is a 16-bit normalized raster map of the elevation of the sampled 3D mesh.
The spatial resolution of color and position samples is user-definable and should be close to the ground sampling distance (GSD) of the source data.
To handle large-scale maps efficiently, the entire pipeline operates in a tiled manner, processing overlapping image patches.
For optimal performance, the sampled area of a patch should correspond to the size of the geospatial 3D map tiles.

After generating a BEV patch, instance segmentation is performed using an appropriate deep learning approach which is explained in detail in Section \ref{sec:results}.
It is important to note that the general framework does not require a specific instance segmentation model but works with any model that is trained to output masks and works with the image patch size.
We allow the specification of occluder classes depending on the model and training data.
For the maritime domain occluders are mostly vehicles, vessels and port infrastructure such as cranes.
This semantic analysis of the rendered color images, identifies areas in the 3D model that require updating, such as occluded or outdated regions.
The predicated 2D masks are then forwarded to two subsequent inpainting stages: a color and a height pass.
In the color pass, deep learning-based generative inpainting techniques remove occluders from the surface texture outputting predicted color images.
Concurrently, the position pass refines the geometric details by inpainting height information, a process closely related to SPIn-NeRF~\cite{mirzaei2023}, and outputting predicted height maps.
This approach leverages the ability of deep-learning based inpainting to generalize well to non-color data, ensuring accurate reconstruction of the scene’s elevation and geometry.

When all image patches have been processed and stored, geometric remeshing and color reprojection are performed.
Remeshing is performed by projecting the 3D vertices of the mesh into the reference frame of the orthogonal camera.
The elevation values of the vertices of the original 3D model are replaced with the inpainted ones. After this step, the occluding geometry conforms to the curvature of the background, which reflects the surrounding environment rather than being strictly flat.
To clean up the projected geometry, vertices are merged based on distance, resulting in a consistent topology.
Color data is then reprojected onto the cleaned vertices, ensuring the updated 3D model is visually accurate.
By default, we generate a second set of texture coordinates with a blending mask which is compatible with the 3D Tiles standard and its underlying GLTF format\cite{gltf}.
Alternatively, a resampling of the original texture using rasterization can be performed to generate a new texture.

Our approach improves upon earlier methods like Engels \textit{et al.} \cite{engels2011facades}, who used a similar concept with 3D point clouds.
However, their approach lacked semantic analysis, resulting in poor segmentation and ineffective reconstruction of geometric structures.
They also required remeshing the 3D point cloud using standard Poisson reconstruction \cite{poisson} and recomputation of the surface textures due to their reliance on point-based plane-fitting for geometry correction.

\section{RESULTS}
\label{sec:results}

Our proposed method for updating 3D geospatial models produces a clean mesh while preserving texture and geometry fidelity of static regions.
Figure \ref{fig:collage_pre_post} shows an overview of the technique on a large scale harbor area.
The results presented here illustrate the effectiveness of the mask-aware inpainting and geometric remeshing pipeline on aerial imagery.

\subsection{Dataset}
\label{sec:dataset}

All aerial data used for the 3D reconstruction of the DSM shown here was captured using a fixed-wing drone flying at an altitude of approximately $\SI{330}{\metre}$.
The area of interest is a port in the south of Bremerhaven, Germany, recorded with a ground sampling distance (GSD) of approximately $\SI{3.7}{\centi\metre}$.  
In total, the area covered by the dataset is roughly $\SI{1}{\kilo\metre}^2$ ($\SI{1030}{\metre} \times \SI{900}{\metre}$).
The generated 3D mesh was sampled from the BEV with a GSD of $\sim\SI{6}{\centi\metre}$ to generate patches with a size of $2048\times2048$ pixels.
We chose this to match the image resolution on which the inpainting methods were trained on.
Additionally, we chose a $50\%$ overlap across all images.
Overlap improves the completeness of detections during instance segmentation by concatenating multiple patches.
In total we processed 195 patches to generate the modified 3D map depicted on the right in Figure \ref{fig:collage_pre_post}.

\subsection{Instance Segmentation and Inpainting}
\label{sec:aiprocessing}

\begin{figure} [ht]
   \begin{center}
   \begin{tabular}{c}
   \includegraphics[height=7cm]{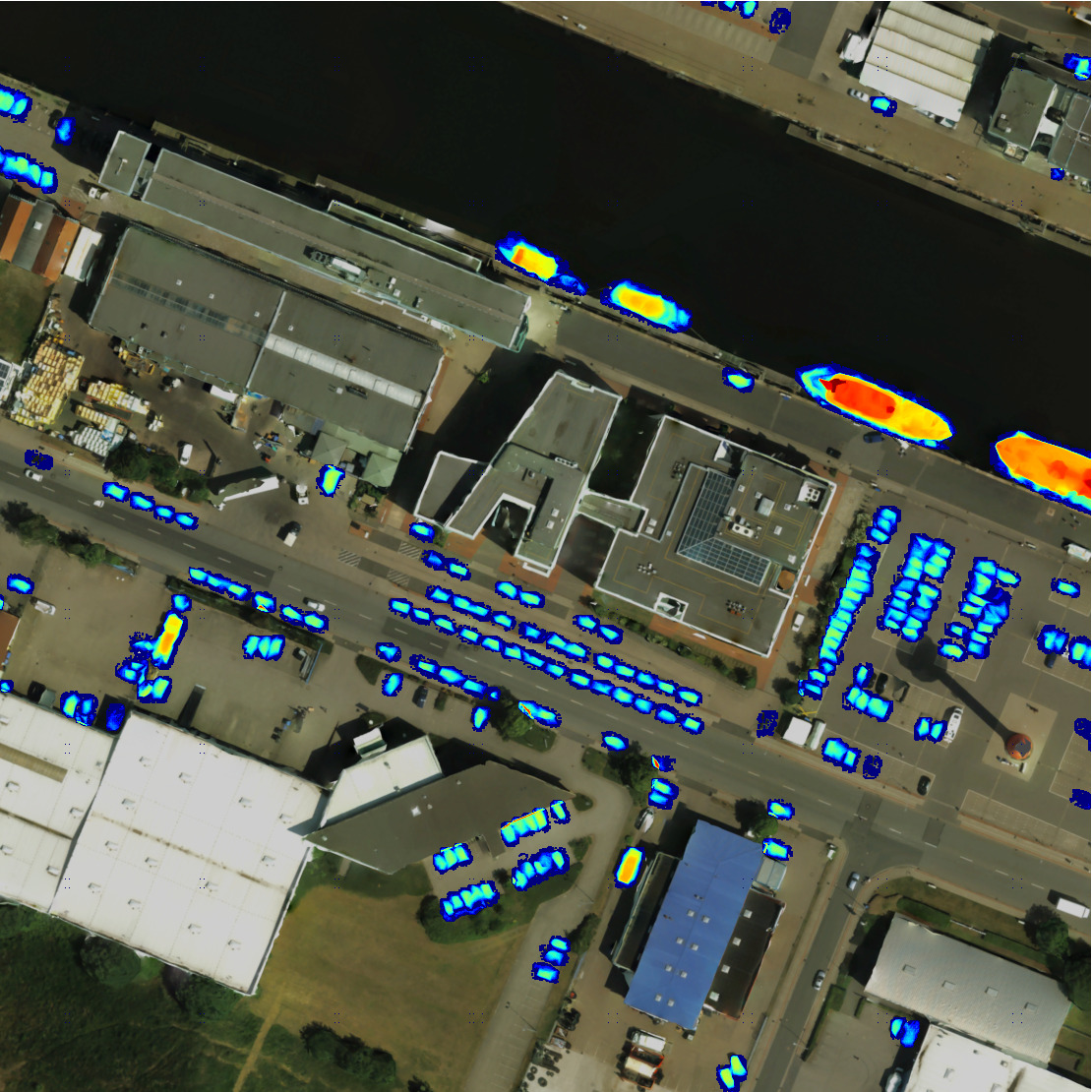}
   \end{tabular}
   \end{center}
   \caption[absdifference] 
   { \label{fig:abs_difference} 
    Heatmap of the normalized distance between source elevation and inpainted elevation for an example patch. The heatmap was overlaid on the source image using the mask generated by the instance segmentation described in this section. False-colored areas indicate regions where the model correctly identified and updated occlusions, demonstrating the accuracy of our mask-aware inpainting approach.}
\end{figure}

Figure \ref{fig:abs_difference} illustrates how we combine instance segmentation and inpainting.
The figure shows the normalized difference between the original (source) elevation and the generated (inpainted) elevation as a heatmap overlaid using the instance masks generated by instance segmentation.
Areas requiring updates are clearly identified by the neural network while all static environments are unaltered.

To accomplish this, we employed YOLOv8 \cite{Jocher_Ultralytics_YOLO_2023}, a state-of-the-art real-time instance segmentation algorithm with the largest configuration (YOLOv8x).
We used pretrained weights on MS COCO \cite{mscoco} and fine-tuned the model by training on the DOTAv2 dataset \cite{dota}, an aerial dataset for instance segmentation that contains the class of interest: vehicle and vessel.
Instance segmentation was performed at LOD6 (the highest resolution) for 195 patches.
The generated masks for processing, such as cars and vessels, were dilated using a $5\times5$ kernel to improve thin masks, and masks from neighboring patches were merged to account for any missing detections at the image edges.
These masks were then downsampled for application to lower LODs.

\begin{figure*} [ht]
   \includegraphics[width=\textwidth]{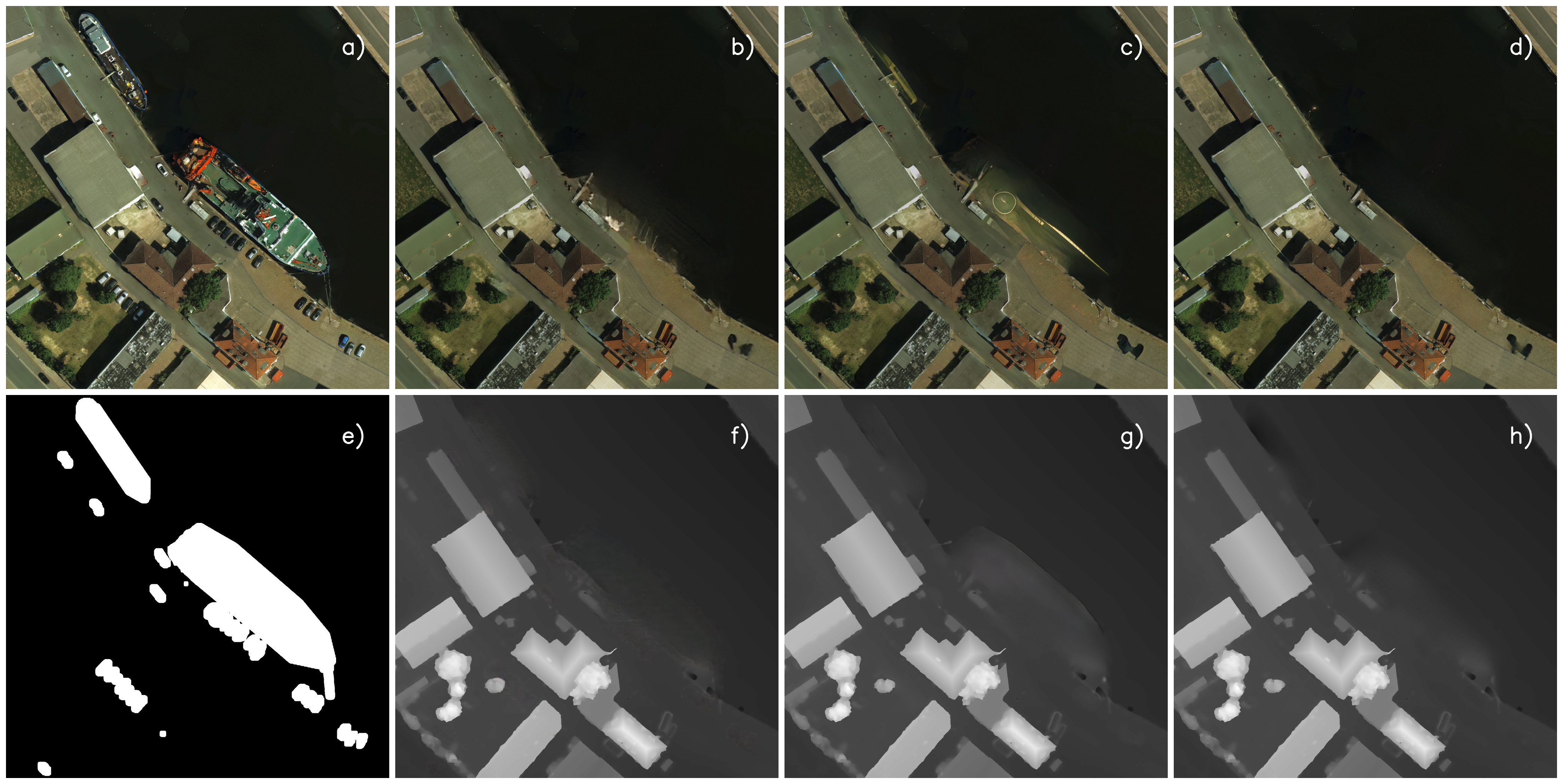}
   \caption[inpainting] 
   { \label{fig:collage_inpainting_techniques} 
     Qualitative comparison of inpainting methods on color and position maps: (a) Source BEV, (b, f) CoModGAN\cite{zhao2021comodgan}, (c, g) MAT\cite{mat2022}, (d, h) LaMa\cite{suvorov2021resolution}, (e) merged and dilated mask from instance segmentation. Note how CoModGAN performs comparable to LaMa for position data, while MAT fails to remove the ship. For color data LaMa outperforms both MAT and CoModGAN for color data leaving only a few shadow artifacts. For quantitative analysis, refer to Table 1.}
\end{figure*} 

\begin{table}[h]
\label{tab:entropy}
\centering
\caption{Mean earth mover distance (EMD) and mean Shannon entropy ($\mathcal{H}$) for color and position data computed over all 195 patches. For $\mathcal{H}$ lower is better, for EMD higher.\\}
\def\arraystretch{1.5}
\adjustbox{max width=\textwidth}{
\begin{tabular}{|c|c|c|c|c|c|c|}
\hline
\multicolumn{1}{|c|}{\multirow{2}{*}{}} & \multicolumn{2}{c|}{\textbf{CoModGAN \cite{zhao2021comodgan}}} & \multicolumn{2}{c|}{\textbf{MAT \cite{mat2022}}} & \multicolumn{2}{c|}{\textbf{LaMa \cite{suvorov2021resolution}}}\\
\cline{2-7}
& Color & Position & Color & Position & Color & Position\\
\hline
\textbf{$\mathcal{H}$} $\downarrow$ & 3.96 & 2.65 & 3.97 & 2.48 & \textbf{3.82} & \textbf{2.26} \\
\hline
\textbf{EMD} $\uparrow$ & 529 & 680 & 469 & 805 & \textbf{649} & \textbf{1072}\\
\hline
\end{tabular}
}
\end{table}

The proposed framework is agnostic to the inpainting method used, so we compared three recent and established architectures: A GAN-based architecture called CoModGAN\cite{zhao2021comodgan}, MAT\cite{mat2022}, a transformer-based architecture and LaMa\cite{suvorov2021resolution}, a neural network architecture using Fourier-based convolutions.
Figure \ref{fig:collage_inpainting_techniques} shows a qualitative comparison of the different inpainting approaches we evaluated for this work.
All three models were trained on the Places dataset\cite{places2}, which includes a wide variety of images with landscape, urban and architectural scenes making it suitable for our use case.
In Table 1 we also provide quantitative metrics to assess the performance of the inpainting methods.

Since we perform evaluation on real-world data, no ground-truth without the occluders is available.
Therefore, mean Shannon entropy and the mean earth mover distance (EMD)~\cite{wasserstein} were used for a comprehensive evaluation.
The removal of occluders effectively means removing high-frequency detail from the image and replacing it with surrounding information.
This directly corresponds to a compression of information which can be measured by a reduction in entropy and change of image statistics.
The EMD measures the cost of transforming the histogram of the inpainted patch to the source histogram, giving a measure of global change.
It is applicable here because the inpainting methods tend to be guided by the image statistics as illustrated in Figure \ref{fig:collage_inpainting_techniques}.

When examining images \textit{d)} and \textit{h)} in Figure \ref{fig:collage_inpainting_techniques} as well as Table 1 it can be seen that LaMa outperforms both MAT and CoModGAN for color as well as position inpainting qualitatively and quantitatively.
Particularly in maintaining consistency with respect to surrounding geometry and texture patterns exemplified by the removal of the berthed vessel.
This is reflected by the reduction in entropy as well as the increase in the EMD.
Especially for roads and parking lots we observed that LaMa also correctly inpainted small details (for example lane markings).
As shown in images \textit{c)} and \textit{g)}, MAT struggles to understand the context of the masked surroundings, often duplicating parts of the image when generating new content or failing to remove structures (exemplified by the vessel hull in the position map, Figure \ref{fig:collage_inpainting_techniques} \textit{g)}).
CoModGAN performed slightly better than MAT for color but not for position data (see Table 1) by producing less smearing on the texture, however it did not respect geometric constraints or color variation (see in Figure \ref{fig:collage_inpainting_techniques} the edge between quay wall and water).
Overall it can be seen that all models perform better on lower-frequency position data rather than color data.
When comparing images \textit{f} through \textit{h} this can be seen when examining how the vessel is inpainted.
MAT and LaMa work on full resolution images, while CoModGAN was limited to $512\times512$ pixels requiring an upsampling in the final stage thus degrading quality.

\subsection{Geometric Remeshing and Projection}
\label{sec:remeshing}

\begin{figure} [ht]
   \begin{center}
   \includegraphics[width=\columnwidth]{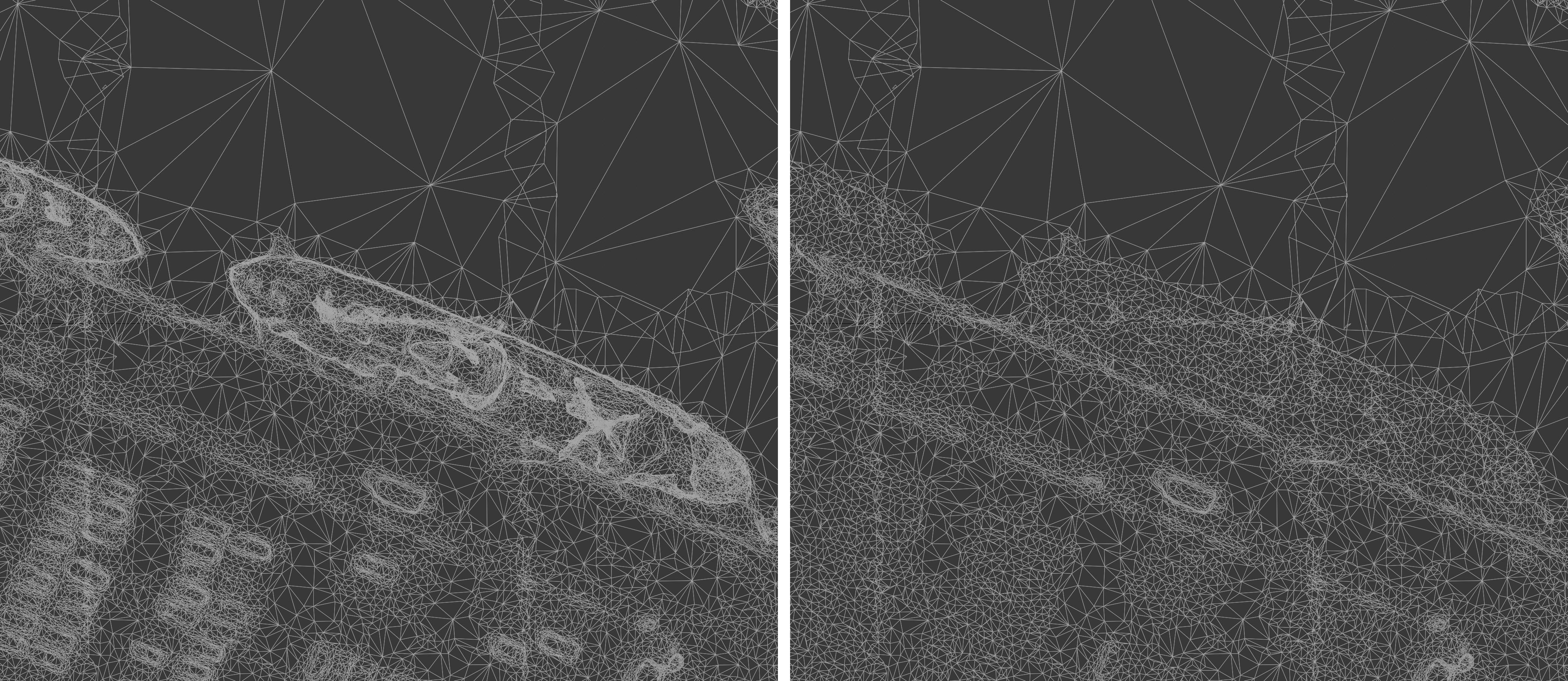}
   \end{center}
   \caption[geometry] 
   { \label{fig:map_large_wire} 
    Wireframe comparison: (Left) Original 3D mesh structure, (Right) Remeshed structure after projection and applying a distance-based merge. While there are remnants of the original mesh in the inpainted parts, the topology is consistent.}
\end{figure} 

Figure \ref{fig:map_large_wire} demonstrates how the inpainted position data is used to alter the geometry of the 3D mesh.
The wireframe overlay (Figure \ref{fig:map_large_wire}, left) shows the original mesh structure, while the right side shows the remeshed structure after applying our method.
Our remeshing technique samples the elevation from the inpainted position map (shown for reference in the center of Figure \ref{fig:map_large_wire}) and effectively addresses overlapping and inconsistent geometric artifacts by merging vertices based on distance, resulting in a consistent topology.
The merge distance is a user-definable parameter and was set to $\SI{0.4}{\metre}$ for the data shown here.
As seen in the figure, the updated geometry better reflects the curvature and contours of the scene, ensuring a more accurate 3D model.
The visible remnants of the original geometry (for example around the berthed vessel) are not artifacts but necessary to allow the remapping of the original texture coordinates to the new model.
By keeping a subset of the original triangle data, it is possible to raster the inpainted color to the original texture.

Our approach not only improves visual accuracy but also addresses compatibility of the updated 3D models with existing geospatial standards like 3D Tiles, aiming to provide seamless integration into modern geospatial applications.

\section{CONCLUSION}
\label{sec:conclusion}

Our proposed method for updating 3D geospatial models in the maritime domain successfully addresses the challenges of occlusion removal and texture fidelity in large-scale remote sensing data.
By employing a combination of state-of-the-art instance segmentation and generative inpainting networks, we maintain both the geometric and visual accuracy of the 3D model and allow alteration without the need for recomputation.
The DSM used for validation provided a testbed, showcasing the effectiveness of our approach across various levels of detail.

The results demonstrate that our method selectively targets dynamic parts of the scene while preserving static environments, thereby minimizing unwanted alterations.
Moreover, the remeshing technique effectively resolves inconsistencies in the geometric structure, leading to a more accurate and visually coherent 3D model.
However, there is room for improvement, particularly in handling artifacts like shadows which all of the inpainting methods fail to remove properly (see Figure \ref{fig:collage_inpainting_techniques}).

Compared to existing methods, our approach offers a unified framework for occlusion removal on 3D mesh data and maintains texture consistency, particularly in complex maritime scenes.
The integration of these techniques ensures that the updated models not only enhance visual fidelity but also maintain compatibility with current geospatial standards, making them suitable for the use in maritime situational awareness and the display of auxiliary information.

Future work should focus on refining instance segmentation, possibly integrating shadow detection techniques, such as those proposed by Wang \textit{et al.}\cite{Wang_2021_CVPR}, or employing specialized architectures for shadow inpainting.
Additionally, training the inpainting network on aerial datasets like iSAID\cite{isaid} and DOTAv2\cite{dota} could enhance its performance in challenging scenarios, including parking lot cells or lane markings which would further improve the quality of the final 3D mesh.


\bibliography{main} 
\bibliographystyle{IEEEtran} 

\end{document}